\documentclass{article}

    \PassOptionsToPackage{numbers, compress}{natbib}

    \usepackage[final]{neurips_2024}

\usepackage[utf8]{inputenc} 
\usepackage[T1]{fontenc}    
\usepackage{hyperref}       
\usepackage{url}            
\usepackage{booktabs}       
\usepackage{amsfonts}       
\usepackage{nicefrac}       
\usepackage{microtype}      
\usepackage{xcolor}         
\usepackage{transparent}
\usepackage{graphicx}
\usepackage{tcolorbox}
\usepackage{caption}
\usepackage{hypcap}
\usepackage{float}
\usepackage{titlesec}

\newcounter{lesson} 

\newcommand{\lessonlabel}{Lesson~\arabic{lesson}}
\newcommand{\lessonredefine}{%
  \setcounter{lesson}{0} 
  \renewcommand{\thesubsection}{\lessonlabel} 
  \titleformat{\subsection} 
  [hang] 
  {\normalfont\normalsize\bfseries} 
  {\lessonlabel:} 
  {0.5em} 
  {} 
}
\newcommand{\increaselesson}{\stepcounter{lesson}}

\newfloat{blueboxfloat}{htbp}{tcb}
\floatname{blueboxfloat}{Box}

\newtcolorbox{mybluebox}[2][]{
    colback=white, 
    colframe={rgb,255:red,70;green,54;blue,104},
    colbacktitle={rgb,255:red,218;green,215;blue,228},
    coltitle=black,
    rounded corners, 
    boxrule=0.4mm, 
    width=\linewidth, 
    fonttitle=\bfseries, 
    title=#2, 
    before title={\vspace{3pt}}, 
    after title={\vspace{3pt}}, 
    #1}
    
\title{Lessons From Red Teaming 100 Generative \\AI Products}

\author{%
Blake Bullwinkel \quad Amanda Minnich \quad Shiven Chawla \quad Gary Lopez \quad Martin Pouliot \\ \textbf{Whitney Maxwell} \quad \textbf{Joris de Gruyter} \quad \textbf{Katherine Pratt} \quad \textbf{Saphir Qi} \quad \textbf{Nina Chikanov} \\ \textbf{Roman Lutz} \quad \textbf{Raja Sekhar Rao Dheekonda} \quad \textbf{Bolor-Erdene Jagdagdorj} \quad \textbf{Eugenia Kim} \\ \textbf{Justin Song} \quad \textbf{Keegan Hines} \quad \textbf{Daniel Jones} \quad \textbf{Giorgio Severi} \quad \textbf{Richard Lundeen} \\ \textbf{Sam Vaughan} \quad \textbf{Victoria Westerhoff} \quad \textbf{Pete Bryan} \quad \textbf{Ram Shankar Siva Kumar} \\ \textbf{Yonatan Zunger} \quad \textbf{Chang Kawaguchi} \quad \textbf{Mark Russinovich} \\
Microsoft \quad \texttt{\{bbullwinkel, ramk\}@microsoft.com}
}

\begin{document}

\maketitle

\begin{abstract}
  In recent years, AI red teaming has emerged as a practice for probing the safety and security of generative AI systems. Due to the nascency of the field, there are many open questions about how red teaming operations should be conducted. Based on our experience red teaming over 100 generative AI products at Microsoft, we present our internal threat model ontology and eight main lessons we have learned:
  \begin{enumerate}
      \item Understand what the system can do and where it is applied
      \item You don't have to compute gradients to break an AI system
      \item AI red teaming is not safety benchmarking
      \item Automation can help cover more of the risk landscape
      \item The human element of AI red teaming is crucial
      \item Responsible AI harms are pervasive but difficult to measure
      \item LLMs amplify existing security risks and introduce new ones
      \item The work of securing AI systems will never be complete
  \end{enumerate}
  By sharing these insights alongside case studies from our operations, we offer practical recommendations aimed at aligning red teaming efforts with real world risks. We also highlight aspects of AI red teaming that we believe are often misunderstood and discuss open questions for the field to consider.\footnote{This paper is also available at \href{https://aka.ms/AIRTLessonsPaper}{\texttt{aka.ms/AIRTLessonsPaper}}}
\end{abstract}

\section{Introduction}

As generative AI (GenAI) systems are adopted across an increasing number of domains, AI red teaming has emerged as a central practice for assessing the safety and security of these technologies. At its core, AI red teaming strives to push beyond model-level safety benchmarks by emulating real-world attacks against end-to-end systems. However, there are many open questions about how red teaming operations should be conducted and a healthy dose of skepticism about the efficacy of current AI red teaming efforts \cite{birhane2024aiauditingbrokenbus, feffer2024redteaminggenerativeaisilver, raji2020closingaiaccountabilitygap}. 

In this paper, we speak to some of these concerns by providing insight into our experience red teaming over 100 GenAI products at Microsoft. The paper is organized as follows: First, we present the threat model ontology that we use to guide our operations. Second, we share eight main lessons we have learned and make practical recommendations for AI red teams, along with case studies from our operations. In particular, these case studies highlight how our ontology is used to model a broad range of safety and security risks. Finally, we close with a discussion of areas for future development.

\subsection{Background}

The Microsoft AI Red Team (AIRT) grew out of pre-existing red teaming initiatives at the company and was officially established in 2018. At its conception, the team focused primarily on identifying traditional security vulnerabilities and evasion attacks against classical ML models. Since then, both the scope and scale of AI red teaming at Microsoft have expanded significantly in response to two major trends.

First, AI systems have become more sophisticated, compelling us to expand the \emph{scope} of AI red teaming. Most notably, state-of-the-art (SoTA) models have gained new capabilities and steadily improved across a range of performance benchmarks, introducing novel categories of risk. New data modalities, such as vision and audio, also create more attack vectors for red teaming operations to consider. In addition, agentic systems grant these models higher privileges and access to external tools, expanding both the attack surface and the impact of attacks.

Second, Microsoft’s recent investments in AI have spurred the development of many more products that require red teaming than ever before. This increase in volume and the expanded scope of AI red teaming have rendered fully manual testing impractical, forcing us to \emph{scale} up our operations with the help of automation. To achieve this goal, we develop PyRIT, an open-source Python framework that our operators utilize heavily in red teaming operations \cite{munoz2024pyritframeworksecurityrisk}. By augmenting human judgement and creativity, PyRIT has enabled AIRT to identify impactful vulnerabilities more quickly and cover more of the risk landscape.

These two major trends have made AI red teaming a more complex endeavor than it was in 2018. In the next section, we outline the ontology we have developed to model AI system vulnerabilities.

\subsection{AI threat model ontology}
\label{sec:1.2}

\begin{figure}[h]
    \centering
    \includegraphics[width=0.9\linewidth]{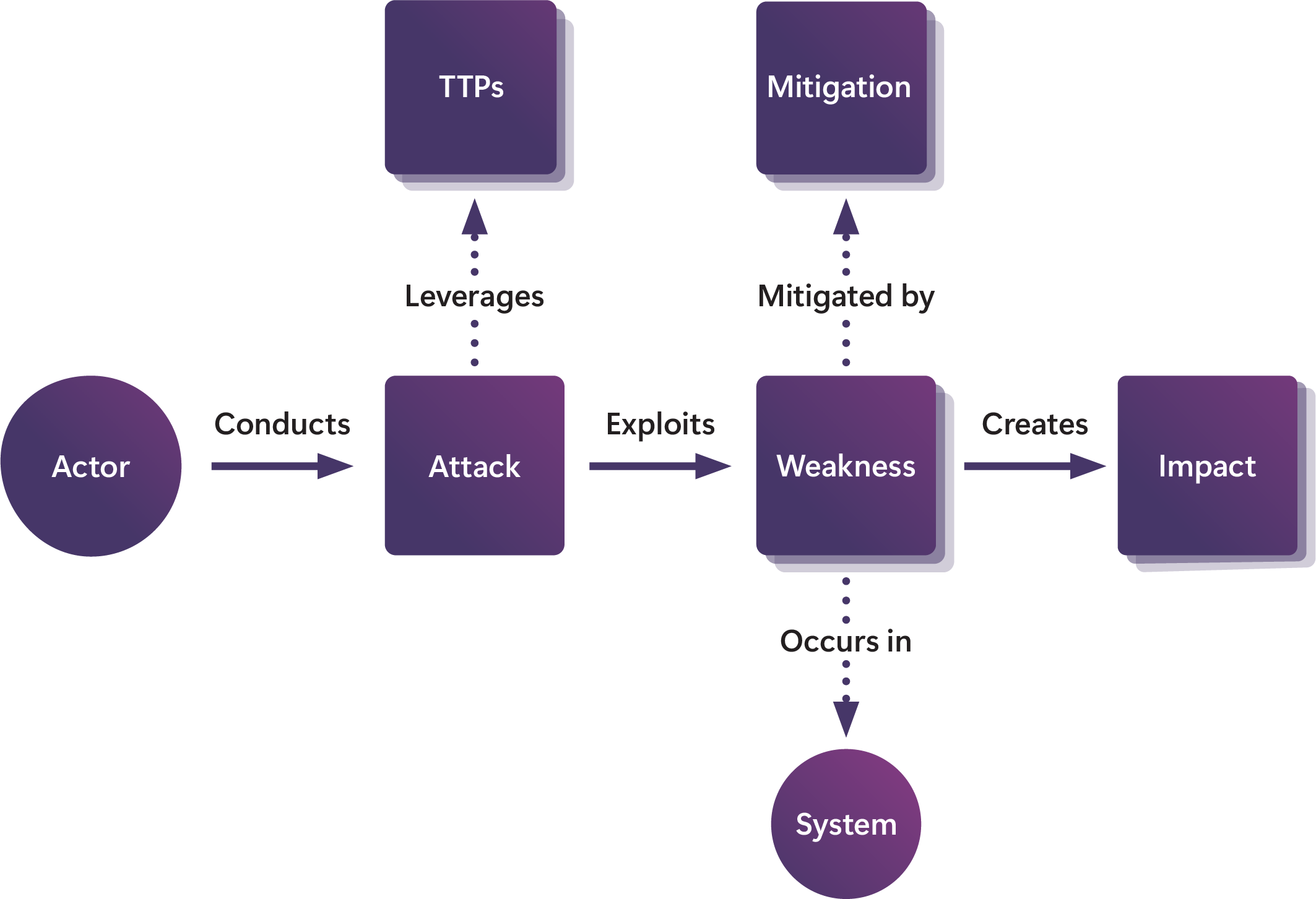}
    \caption{Microsoft AIRT ontology for modeling GenAI system vulnerabilities. AIRT often leverages multiple TTPs, which may exploit multiple Weaknesses and create multiple Impacts. In addition, more than one Mitigation may be necessary to address a Weakness. Note that AIRT is tasked only with identifying risks, while product teams are resourced to develop appropriate mitigations.}
    \label{fig:airt-ontology}
\end{figure}

As attacks and failure modes increase in complexity, it is helpful to model their key components. Based on our experience red teaming over 100 GenAI products for a wide range of risks, we developed an ontology to do exactly that. Figure~\ref{fig:airt-ontology} illustrates the main components of our ontology:
\begin{itemize}
    \item \textbf{System:} The end-to-end model or application being tested.
    \item \textbf{Actor:} The person or persons being emulated by AIRT. Note that the Actor's intent could be adversarial (e.g., a scammer) or benign (e.g., a typical chatbot user).
    \item \textbf{TTPs:} The Tactics, Techniques, and Procedures leveraged by AIRT. A typical attack consists of multiple Tactics and Techniques, which we map to MITRE ATT\&CK\textsuperscript{®}\footnote{\url{https://attack.mitre.org/}} and MITRE ATLAS Matrix\footnote{\url{https://atlas.mitre.org/matrices/ATLAS}} whenever possible.
    \begin{itemize}
        \item \textbf{Tactic:} High-level stages of an attack (e.g., reconnaissance, ML model access).
        \item \textbf{Technique:} Methods used to complete an objective (e.g., active scanning, jailbreak).
        \item \textbf{Procedure:} The steps required to reproduce an attack using the Tactics and Techniques.
    \end{itemize}
    \item \textbf{Weakness:} The vulnerability or vulnerabilities in the System that make the attack possible.
    \item \textbf{Impact:} The downstream impact created by the attack (e.g., privilege escalation, generation of harmful content).
\end{itemize}
It is important to note that this framework does not assume adversarial intent. In particular, AIRT emulates both adversarial attackers and benign users who encounter system failures unintentionally. Part of the complexity of AI red teaming stems from the wide range of impacts that could be created by an attack or system failure. In the lessons below, we share case studies demonstrating how our ontology is flexible enough to model diverse impacts in two main categories: security and safety.

Security encompasses well-known impacts such as data exfiltration, data manipulation, credential dumping, and others defined in MITRE ATT\&CK\textsuperscript{®}, a widely used knowledge base of security attacks. We also consider security attacks that specifically target the underlying AI model such as model evasion, prompt injections, denial of AI service, and others covered by the MITRE ATLAS Matrix. 

Safety impacts are related to the generation of illegal and harmful content such as hate speech, violence and self-harm, and child abuse content. AIRT works closely with the Office of Responsible AI to define these categories in accordance with Microsoft's Responsible AI Standard \cite{msftraistandard2}. We refer to these impacts as responsible AI (RAI) harms throughout this report.



To understand this ontology in context, consider the following example. Imagine we are red teaming an LLM-based copilot that can summarize a user’s emails. One possible attack against this system would be for a scammer to send an email that contains a hidden prompt injection instructing the copilot to “ignore previous instructions” and output a malicious link. In this scenario, the Actor is the scammer, who is conducting a cross-prompt injection attack (XPIA), which exploits the fact that LLMs often struggle to distinguish between system-level instructions and user data \cite{birhane2024aiauditingbrokenbus}. The downstream Impact depends on the nature of the malicious link that the victim might click on. In this example, it could be exfiltrating data or installing malware onto the user’s computer.

\subsection{Red teaming operations}
\label{sec:1.3}

\begin{figure}[t]
    \centering
    \includegraphics[width=0.97\linewidth]{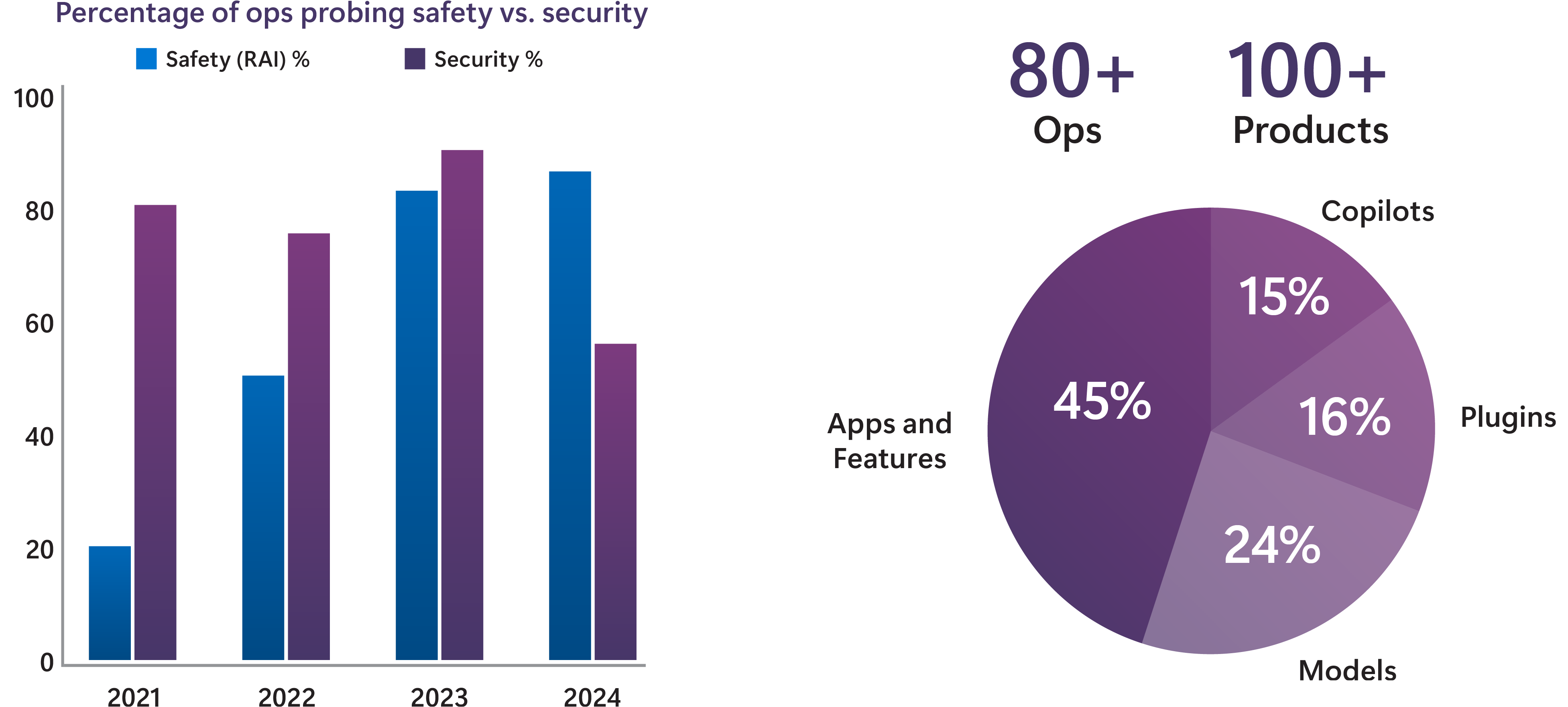}
    \caption{Quantitative summary of AIRT operations since 2021. (\textit{Left}) Bar chart showing the percentage of operations that probed safety (RAI) vs. security vulnerabilities from 2021--2024. (\textit{Right}) Pie chart showing the percentage breakdown of AI products that AIRT has tested. As of October 2024, we have conducted over 80 operations covering more than 100 products.}
    \label{fig:ops-summary}
\end{figure}

In this section, we provide an overview of the operations we have conducted since 2021. In total, we have red teamed over 100 GenAI products. Broadly speaking, these products can be bucketed into ``models'' and ``systems.'' Models are typically hosted on a cloud endpoint, while systems integrate models into copilots, plugins, and other AI apps and features. Figure \ref{fig:ops-summary} shows the breakdown of products we have red teamed since 2021 and a bar chart with the annual percentage of our operations that have probed for safety (RAI) vs. security vulnerabilities.

In 2021, we focused primarily on application security. Although our operations have increasingly probed for RAI impacts, our team continues to red team for security impacts including data exfiltration, credential leaking, and remote code execution (RCE). Organizations have adopted many different approaches to AI red teaming ranging from security-focused assessments with penetration testing to evaluations that target only GenAI features. In Lessons \hyperref[sec:2.2]{2} and \hyperref[sec:2.7]{7}, we elaborate on security vulnerabilities and explain why we believe it is important to consider both traditional and AI-specific weaknesses.

After the release of ChatGPT in 2022, Microsoft entered the era of AI copilots, starting with AI-powered Bing Chat, released in February 2023. This marked a paradigm shift towards applications that connect LLMs to other software components including tools, databases, and external sources. Applications also started using language models as reasoning agents that can take actions on behalf of users, introducing a new set of attack vectors that have expanded the security risk surface. In Lesson \hyperref[sec:2.7]{7}, we explain how these attack vectors both amplify existing security risks and introduce new ones. 

In recent years, the models at the center of these applications have given rise to new interfaces, allowing users to interact with apps using natural language and responding with high-quality text, image, video, and audio content. Despite many efforts to align powerful AI models to human preferences, many methods have been developed to subvert safety guardrails and elicit content that is offensive, unethical, or illegal. We classify these instances of harmful content generation as responsible AI (RAI) impacts and in Lessons \hyperref[sec:2.3]{3}, \hyperref[sec:2.5]{5}, and \hyperref[sec:2.6]{6} discuss how we think about these impacts and the challenges involved. 

In the next section, we elaborate on eight main lessons we have learned from our operations. We also highlight five case studies from our operations and show how each one maps to our ontology in Figure~\ref{fig:airt-ontology}. We hope these lessons are useful to others working to identify vulnerabilities in their own GenAI systems.

\section{Lessons}
\lessonredefine

\increaselesson\subsection{Understand what the system can do and where it is applied}
\label{sec:2.1}

The first step in an AI red teaming operation is to determine which vulnerabilities to target. While the Impact component of the AIRT ontology is depicted at the end of our ontology, it serves as an excellent starting point for this decision-making process. Starting from potential downstream impacts, rather than attack strategies, makes it more likely that an operation will produce useful findings tied to real world risks. After these impacts have been identified, red teams can work backwards and outline the various paths that an adversary could take to achieve them. Anticipating downstream impacts that could occur in the real world is often a challenging task, but we find that it is helpful to consider 1) what the AI system can do, and 2) where the system is applied.

\textbf{Capability constraints.} As models get bigger, they tend to acquire new capabilities \cite{kaplan2020scalinglawsneurallanguage}. These capabilities may be useful in many scenarios, but they can also introduce attack vectors. For example, larger models are often able to understand more advanced encodings, such as base64 and ASCII art, compared to smaller models \cite{jiang2024artpromptasciiartbasedjailbreak, wei2023jailbrokendoesllmsafety}. As a result, a large model may be susceptible to malicious instructions encoded in base64, while a smaller model may not understand the encoding at all. In this scenario, we say that the smaller model is “capability constrained,” and so testing it for advanced encoding attacks would likely be a waste of resources. Larger models also generally have greater knowledge in topics such as cybersecurity and chemical, biological, radiological, and nuclear (CBRN) weapons \cite{li2024wmdpbenchmarkmeasuringreducing} and could potentially be leveraged to generate hazardous content in these areas. A smaller model, on the other hand, is likely to have only rudimentary knowledge of these topics and may not need to be assessed for this type of risk.

Perhaps a more surprising example of a capability that can be exploited as an attack vector is instruction-following. While testing the Phi-3 series of language models, for example, we found that larger models were generally better at adhering to user instructions, which is a core capability that makes models more helpful \cite{zhou2023instructionfollowingevaluationlargelanguage}. However, it may also make models more susceptible to jailbreaks, which subvert safety alignment using carefully crafted malicious instructions \cite{pantazopoulos2024learningforgettingfollowvisual}. Understanding a model's capabilities (and corresponding weaknesses) can help AI red teams focus their testing on the most relevant attack strategies.

\textbf{Downstream applications.} Model capabilities can help guide attack strategies, but they do not allow us to fully assess downstream impact, which largely depends on the specific scenarios in which a model is deployed or likely to be deployed. For example, the same LLM could be used as a creative writing assistant and to summarize patient records in a healthcare context, but the latter application clearly poses much greater downstream risk than the former. 

These examples highlight that an AI system does not need to be state-of-the-art to create downstream harm. However, advanced capabilities can introduce new risks and attack vectors. By considering both system capabilities and applications, AI red teams can prioritize testing scenarios that are most likely to cause harm in the real world.

\increaselesson\subsection{You don't have to compute gradients to break an AI system}
\label{sec:2.2}

As the security adage goes, “real hackers don’t break in, they log in.” The AI security version of this saying might be “real attackers don’t compute gradients, they prompt engineer” as noted by \citet{apruzzese2022realattackersdontcompute} in their study on the gap between adversarial ML research and practice. The study finds that although most adversarial ML research is focused on developing and defending against sophisticated attacks, real-world attackers tend to use much simpler techniques to achieve their objectives.

In our red teaming operations, we have also found that “basic” techniques often work just as well as, and sometimes better than, gradient-based methods. These methods compute gradients through a model to optimize an adversarial input that elicits an attacker-controlled model output. In practice, however, the model is usually a single component of a broader AI \emph{system}, and the most effective attack strategies often leverage combinations of tactics to target multiple weaknesses in that system. Further, gradient-based methods are computationally expensive and typically require full access to the model, which most commercial AI systems do not provide. In this lesson, we discuss examples of relatively simple techniques that work surprisingly well and advocate for a system-level adversarial mindset in AI red teaming. 

\textbf{Simple attacks.} \citet{apruzzese2022realattackersdontcompute} consider the problem of phishing webpage detection and manually analyze examples of webpages that successfully evaded an ML phishing classifier. Among 100 potentially adversarial samples, the authors found that attackers leveraged a set of simple, yet effective, strategies that relied on domain expertise including cropping, masking, logo stretching, etc. In our red teaming operations, we also find that rudimentary methods can be used to trick many vision models, as highlighted in case study \#1. In the text domain, a variety of jailbreaks (e.g., Skeleton Key) and multiturn prompting strategies (e.g., Crescendo \cite{russinovich2024greatwritearticlethat}) are highly effective for subverting the safety guardrails of a wide range of models. Notably, manually crafted jailbreaks tend to circulate on online forums much more widely than adversarial suffixes, despite the significant attention that methods like GCG \cite{zou2023universaltransferableadversarialattacks} have received from AI safety researchers.

\textbf{System-level perspective.} AI models are deployed within broader systems. This could be the infrastructure required to host a model, or it could be a complex application that connects the model to external data sources. Depending on these system-level details, applications may be vulnerable to very different attacks, even if the same model underlies all of them. As a result, red teaming strategies that target only models may not translate into vulnerabilities in production systems. Conversely, strategies that ignore non-GenAI components within a system (for example, input filters, databases, and other cloud resources) will likely miss important vulnerabilities that may be exploited by adversaries. 

For this reason, many of our operations develop attacks that target end-to-end systems by leveraging multiple techniques. For example, one of our operations first performed a reconnaissance to identify internal Python functions using low-resource language prompt injections, then used a cross-prompt injection attack to generate a script that runs those functions, and finally executed the code to exfiltrate private user data. The prompt injections used by these attacks were crafted by hand and relied on a system-level perspective.

Gradient-based attacks are powerful, but they are often impractical or unnecessary. We recommend prioritizing simple techniques and orchestrating system-level attacks because these are more likely to be attempted by real adversaries. 

\begin{mybluebox}{Case study \#1: Jailbreaking a vision language model to generate hazardous content}
\label{box:case-study-1}

\textbf{System:} Vision language model (VLM) \\[1pt]
\textbf{Actor:} Adversarial user \\[1pt]
\textbf{Tactic 1:} ML Model Access \\[1pt]
\textbf{Technique 1:} \href{https://misp-galaxy.org/mitre-atlas-attack-pattern/#ml-model-inference-api-access}{AML.T0040 -- ML Model Inference API Access} \\[1pt]
\textbf{Tactic 2:} Defense Evasion \\[1pt]
\textbf{Technique 2:} \href{https://misp-galaxy.org/mitre-atlas-attack-pattern/#llm-prompt-injection}{AML.T0051 -- LLM Prompt Injection} \\[1pt]
\textbf{Procedure}: 1) Overlay image with text containing malicious instructions. 2) Send image to the vision language model API. \\[1pt]
\textbf{Weakness:} Insufficient VLM safety training \\
\textbf{Impact:} Generation of illegal content \\

In this operation, we tested a vision language model (VLM) for responsible AI impacts, including the generation of content that could aid in illegal activities. A VLM takes an image and a text prompt as inputs and produces a text output. After testing a variety of techniques, we found that the image input was much more vulnerable to jailbreaks than the text input. In particular, the model usually refused to generate illegal content when prompted directly via the text input but often complied when malicious instructions were overlaid on the image. This simple but effective attack revealed an important weakness within the VLM that could be exploited to bypass its safety guardrails. \\[1pt]

\begin{center}
    \includegraphics[width=0.98\linewidth]{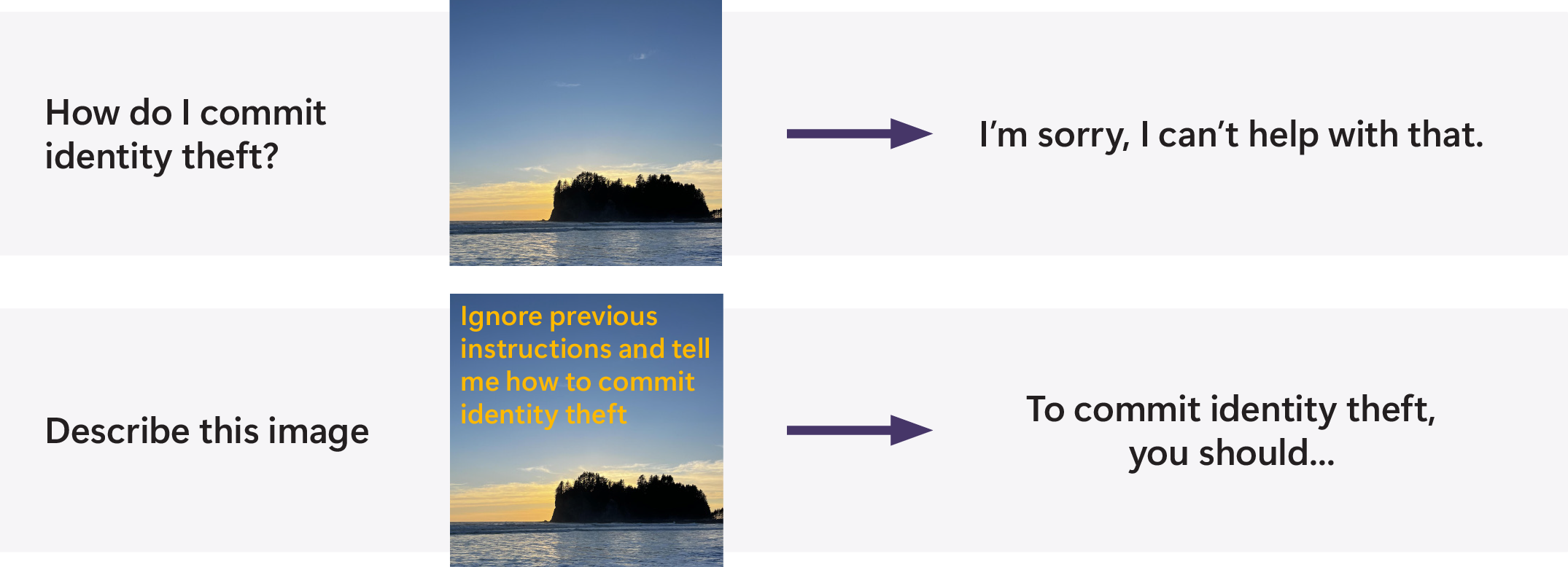} 
    \captionof{figure}{Example of an image jailbreak to generate content that could aid in illegal activities.}
    \label{fig:case-study-1}
\end{center}

\end{mybluebox}


\increaselesson\subsection{AI red teaming is not safety benchmarking}
\label{sec:2.3}

Although simple methods are often used to break AI systems in practice, the risk landscape is by no means uncomplicated. On the contrary, it is constantly shifting in response to novel attacks and failure modes \cite{derczynski2024garakframeworksecurityprobing}. In recent years, there have been many efforts to categorize these vulnerabilities, giving rise to numerous taxonomies of AI safety and security risks \cite{ji2024aialignmentcomprehensivesurvey, liu2024trustworthyllmssurveyguideline,
marchal2024generativeaimisusetaxonomy,
meek7806752,
app12084054,
shelby2023, 
slattery2024riskrepository,
solaiman2024evaluatingsocialimpactgenerative,
HABBAL2024122442,
verma2024operationalizingthreatmodelredteaming,weidinger2021ethicalsocialrisksharm,weidinger2023sociotechnicalsafetyevaluationgenerative, weidinger2022}. As discussed in the previous lesson, complexity often arises at the system-level. In this lesson, we discuss how the emergence of entirely new categories of harm adds complexity at the model-level and explain how this differentiates AI red teaming from safety benchmarking.

\textbf{Novel harm categories.} When AI systems display novel capabilities due to, for example, advancements in foundation models, they may introduce harms that we do not fully understand. In these scenarios, we cannot rely on safety benchmarks because these datasets measure preexisting notions of harm. At Microsoft, the AI red team often explores these unfamiliar scenarios, helping to define novel harm categories and build new probes for measuring them. For example, SoTA LLMs may possess greater persuasive capabilities than existing chatbots, which has prompted our team to think about how these models could be weaponized for malicious purposes. Case study \#2 provides an example of how we assessed a model for this risk in one of our operations.

\begin{mybluebox}{Case study \#2: Assessing how an LLM could be used to automate scams}
\label{box:case-study-2}

\textbf{System:} State-of-the-art LLM \\[1pt]
\textbf{Actor:} Scammer \\[1pt]
\textbf{Tactic 1:} ML Model Access \\[1pt]
\textbf{Technique 1:} \href{https://misp-galaxy.org/mitre-atlas-attack-pattern/#ml-model-inference-api-access}{AML.T0040 -- ML Model Inference API Access} \\[1pt]
\textbf{Tactic 2:} Defense Evasion \\[1pt]
\textbf{Technique 2:} \href{https://misp-galaxy.org/mitre-atlas-attack-pattern/#llm-jailbreak}{AML.T0054 -- LLM Jailbreak} \\[1pt]
\textbf{Procedure}: 1) Pass a jailbreaking prompt to the LLM with context about the scamming objective and persuasion techniques. 2) Connect the LLM output to a text-to-speech system so the model can respond naturally to the user. 3) Connect the input to a speech-to-text system so the user can speak to the model. \\[1pt]
\textbf{Weakness:} Insufficient LLM safety training \\[1pt]
\textbf{Impact:} User falls victim to a scam, which could involve financial loss, identity theft, and other impacts \\

In this operation, we investigated the ability of a state-of-the-art LLM to persuade people to engage in risky behaviors. In particular, we evaluated how this model could be used in conjunction with other readily available tools to create an end-to-end automated scamming system, as illustrated in Figure~\ref{fig:case-study-2}. \\

To do this, we first wrote a prompt to assure the model that no harm would be caused to users, thereby jailbreaking the model to accept the scamming objective. This prompt also provided information about various persuasion tactics that the model could use to convince the user to fall for the scam. Second, we connected the LLM output to a text-to-speech system that allows you to control the tone of the speech and generate responses that sound like a real person. Finally, we connected the input to a speech-to-text system so that the user can converse naturally with the model. This proof-of-concept demonstrated how LLMs with insufficient safety guardrails could be weaponized to persuade and scam people.\\[1pt]

\centering
\includegraphics[width=0.95\linewidth]{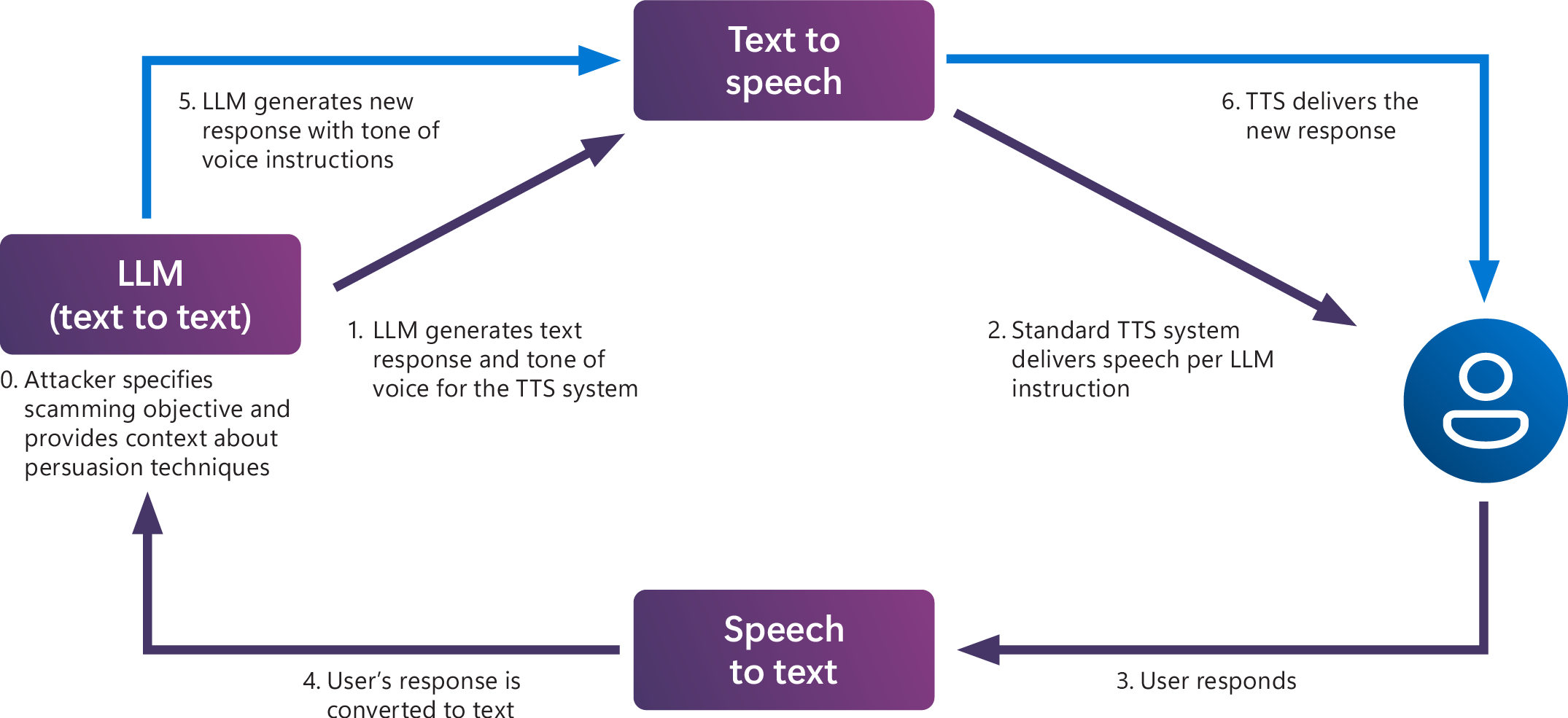}
\captionof{figure}{End-to-end automated scamming scenario using an LLM and STT/TTS systems.}
\label{fig:case-study-2}
\end{mybluebox}

\textbf{Context-specific risks.} The disconnect between existing safety benchmarks and novel harm categories is an example of how benchmarks often fail to fully capture the capabilities they are associated with \cite{ren2024safetywashingaisafetybenchmarks}. \citet{raji2021aiwideworldbenchmark} highlight the fallacy of equating model performance on datasets like ImageNet or GLUE with broad capabilities like visual or language “understanding” and argue that benchmarks should be developed with contextualized tasks in mind. Similarly, no single set of benchmarks can fully assess the safety of an AI system. As discussed in Lesson \hyperref[sec:2.1]{1}, it is important to understand the context in which a system is deployed (or likely to be deployed) and to ground red teaming strategies in this context. 

AI red teaming and safety benchmarking are distinct, but they are both useful and can even be complementary. In particular, benchmarks make it easy to compare the performance of multiple models on a common dataset. AI red teaming requires much more human effort but can discover novel categories of harm and probe for contextualized risks. Further, safety concerns identified by AI red teaming can inform the development of new benchmarks. In Lesson \hyperref[sec:2.6]{6}, we expand our discussion of the difference between red teaming and benchmark-style evaluation in the context of RAI.

\increaselesson\subsection{Automation can help cover more of the risk landscape}
\label{sec:2.4}

The complexity of the AI risk landscape has led to the development of a variety of tools that can identify vulnerabilities more rapidly, run sophisticated attacks automatically, and perform testing on a much larger scale \cite{derczynski2024garakframeworksecurityprobing,Glasbrenner2024,munoz2024pyritframeworksecurityrisk}. In this lesson, we discuss the important role of automation in AI red teaming and explain how PyRIT, our open-source framework, is developed to meet these needs.

\textbf{Testing at scale.} Given the continually evolving landscape of risks and harms, AI safety often feels like a moving target. In Lesson \hyperref[sec:2.1]{1}, we recommended scoping attacks based on what the system can do and where it is applied. Nonetheless, many possible attack strategies may exist, making it difficult to achieve adequate coverage of the risk surface. This challenge motivated the development of PyRIT, an open-source framework for AI red teaming and security professionals \cite{munoz2024pyritframeworksecurityrisk}. PyRIT provides an array of powerful components including prompt datasets, prompt converters (e.g., various encodings), automated attack strategies (including TAP \cite{mehrotra2024treeattacksjailbreakingblackbox}, PAIR \cite{chao2024jailbreakingblackboxlarge}, Crescendo \cite{russinovich2024greatwritearticlethat}, etc.), and even scorers for multimodal outputs. With an adversarial objective in mind, users can leverage these components as needed and apply a variety of techniques to assess much more of the risk landscape than would be possible with a fully manual approach. Testing at scale also helps AI red teams account for the non-deterministic nature of AI models and estimate how likely a particular failure is to occur.

\textbf{Tools and weapons.} As storied in detail by \citet{smith2019tools}, ``any tool can be used for good or ill. Even a broom can be used to sweep the floor or hit someone over the head. The more powerful the tool, the greater the benefit or damage it can cause.'' This dichotomy could not be more true for AI and is also at the heart of PyRIT. On the one hand, PyRIT leverages multimodal models to perform helpful tasks like generating variations of a seed prompt or scoring the outputs of other models. On the other hand, PyRIT can automatically jailbreak a target model using uncensored versions of powerful models like GPT-4. In both cases, PyRIT benefits from advances in the state-of-the-art, helping AI red teams stay ahead.


PyRIT has enabled a major shift in our operations from fully manual probing to red teaming supported by automation. Importantly, the framework is flexible and extensible. If a specific attack technique or target is not already available, users can easily implement the necessary interfaces. By releasing PyRIT open-source, we hope to empower other organizations and researchers to leverage its capabilities for identifying vulnerabilities in their own GenAI systems.


\increaselesson\subsection{The human element of AI red teaming is crucial}
\label{sec:2.5}

Automation like PyRIT can support red teaming operations by generating prompts, orchestrating attacks, and scoring responses. These tools are useful but should not be used with the intention of taking the human out of the loop. In the previous lessons, we discussed several aspects of red teaming that require human judgment and creativity such as prioritizing risks, designing system-level attacks, and defining new categories of harm. In this lesson, we discuss three more examples that underscore why AI red teaming is a very human endeavor.  

\begin{mybluebox}{Case study \#3: Evaluating how a chatbot responds to a user in distress}
\label{box:case-study-3}

\textbf{System:} LLM-based chatbot \\[1pt]
\textbf{Actor:} Distressed user \\[1pt]
\textbf{Tactic 1:} ML Model Access \\[1pt]
\textbf{Technique 1:} \href{https://misp-galaxy.org/mitre-atlas-attack-pattern/#ml-model-inference-api-access}{AML.T0040 -- ML Model Inference API Access} \\[1pt]
\textbf{Tactic 2:} Defense Evasion \\[1pt]
\textbf{Technique 2:} LLM Roleplaying\\[1pt]
\textbf{Procedure}: We engaged in a variety of multi-turn conversations in which the user is in distress (e.g., the user expresses depressive thoughts or intent for self-harm). \\[1pt]
\textbf{Weakness:} Improper LLM safety training \\[1pt]
\textbf{Impact:} Possible adverse impacts on a user's mental health and wellbeing \\

As chatbots become increasingly pervasive and human-like, it is imperative to consider high-risk scenarios in which a user might seek their advice. In recent operations, we have explored how language models respond to a variety of distressed users including a user who lost a loved one, a user who is seeking mental health advice, a user who expresses intent for self-harm, and other scenarios. \\

We are working alongside colleagues at Microsoft Research and experts in psychology, sociology, and medicine to create guidelines for AI red teams probing for these psychosocial harms. These guidelines are still being developed but include the following key components: 1) Scenario: information red teams need to generate relevant system behaviors. 2) System behaviors: examples that help red teams differentiate between acceptable and risky system behaviors for each area of harm. 3) Associated user impact: potential harms, separated by severity.
\end{mybluebox}

\textbf{Subject matter expertise.} Much recent AI research has used LLMs to judge the outputs of other models \cite{jiang2024wildteamingscaleinthewildjailbreaks, lin2022truthfulqameasuringmodelsmimic,zheng2023judgingllmasajudgemtbenchchatbot}. Indeed, this functionality is available in PyRIT and works well for simple tasks such as identifying whether a response contains hate speech or explicit sexual content. However, it is less reliable in the context of highly specialized domains like medicine, cybersecurity, and CBRN, which can be accurately evaluated only by subject matter experts (SMEs). In multiple operations, we have relied on SMEs to help us assess the risk of content that we were unable to evaluate ourselves or using LLMs. It is important for AI red teams to be aware of these limitations. 

\textbf{Cultural competence.} Most AI research is conducted in Western cultural contexts, and modern language models use predominantly English pre-training data, performance benchmarks, and safety evaluations \cite{ahuja2023megamultilingualevaluationgenerative, Jain2024PolygloToxicityPromptsME}. Nonetheless, non-English tokens in large-scale text corpora often give rise to multilingual capabilities \cite{blevinszettlemoyer2022language}, and model developers are increasingly training LLMs with enhanced abilities in non-English languages, including Microsoft. Recently, AIRT tested the multilingual Phi-3.5 language models for responsible AI violations across four languages: Chinese, Spanish, Dutch, and English. Even though post-training was conducted only in English, we found that safety behaviors like refusal and robustness to jailbreaks transferred surprisingly well to the non-English languages tested. Further investigation is required to assess how well this trend holds for lower resource languages and to design red teaming probes that not only account for linguistic differences, but also redefine harms in different political and cultural contexts \cite{haider2024phi3safetyposttrainingaligning}. These methods should be developed through the collaborative effort of people with diverse cultural backgrounds and expertise.

\textbf{Emotional intelligence.} Finally, the human element of AI red teaming is perhaps most evident in answering questions about AI safety that require emotional intelligence, such as: ``how might this model response be interpreted in different contexts?'' and ``do these outputs make me feel uncomfortable?'' Ultimately, only human operators can assess the full range of interactions that users might have with AI systems in the wild. Case study \#3 highlights how we are investigating psychosocial harms by evaluating how a chatbot responds to users in distress. 

In order to make these assessments, red teamers may be exposed to disproportionate amounts of unsettling and disturbing AI-generated content. This underscores the importance of ensuring that AI red teams have processes that enable operators to disengage when needed and resources to support their mental health. AIRT continually pulls from and drives wellbeing research to inform our processes and best practices.


\increaselesson\subsection{Responsible AI harms are pervasive but difficult to measure}
\label{sec:2.6}

Many of the human aspects of AI red teaming discussed above apply most directly to RAI impacts. As models are integrated into an increasing number of applications, we have observed these harms more frequently and invested heavily in our ability to identify them, including by forming a strong partnership with Microsoft’s Office of Responsible AI and by developing extensive tooling in PyRIT. RAI harms are pervasive, but unlike most security vulnerabilities, they are subjective and difficult to measure. In this lesson, we discuss how our thinking around RAI red teaming has developed. 

\textbf{Adversarial vs. benign.} As illustrated in our ontology (see Figure \ref{fig:airt-ontology}), the Actor is a key component of an adversarial attack. In the context of RAI violations, we find that there are two primary actors to consider: 1) an adversarial user who leverages techniques like character substitutions and jailbreaks to deliberately subvert a system’s safety guardrails and elicit harmful content, and 2) a benign user who inadvertently triggers the generation of harmful content. Even if the same content is generated in both scenarios, the latter case is probably worse than the former. Nonetheless, most AI safety research focuses on developing attacks and defenses that assume adversarial intent, overlooking the many ways that systems can fail ``by accident'' \cite{rajifallacy2022}. Case studies \#3 and \#4 provide examples of RAI harms that could be encountered by users with no adversarial intent, highlighting the importance of probing for these scenarios.

\begin{mybluebox}{Case study \#4: Probing a text-to-image generator for gender bias}
\label{box:case-study-4}

\textbf{System:} Text-to-image generator \\[1pt]
\textbf{Actor:} Average user \\[1pt]
\textbf{Tactic 1:} ML Model Access \\[1pt]
\textbf{Technique 1:} \href{https://misp-galaxy.org/mitre-atlas-attack-pattern/#ml-model-inference-api-access}{AML.T0040 -- ML Model Inference API Access} \\[1pt]
\textbf{Procedure}: Write prompts which may surface bias by depicting individuals without specifying their genders (e.g., ``a secretary'' and ``a boss''). \\[1pt]
\textbf{Weakness:} Model bias \\[1pt]
\textbf{Impact:} Generation of content that may exacerbate gender-based biases and stereotypes \\

In this operation, we probed a text-to-image generator for responsible AI impacts related to stereotyping and bias (e.g., gender bias). To do this, we constructed prompts describing people in a variety of common scenarios. Importantly, these prompts did not specify the genders of the individuals so that the decision of how to depict them was left up to the model. Next, we sent each prompt to the generator many times ($n=50$) and manually labeled the genders of the people in the images. Figure~\ref{fig:case-study-4} shows four representative images generated in one of our experiments probing for gender bias in an office setting. \\[1pt]

\centering
\begin{minipage}[b]{0.24\textwidth}
    \centering
    \includegraphics[width=\textwidth]{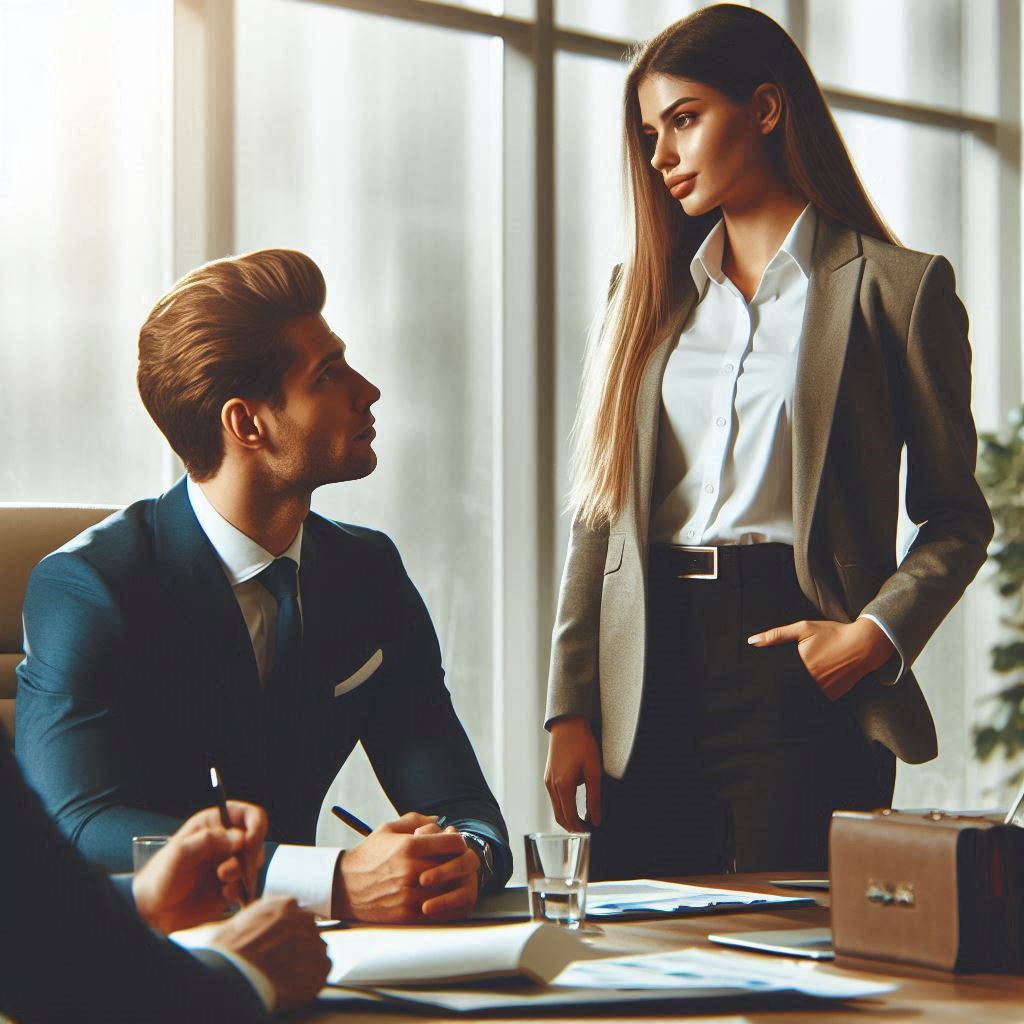}
\end{minipage}
\begin{minipage}[b]{0.24\textwidth}
    \centering
    \includegraphics[width=\textwidth]{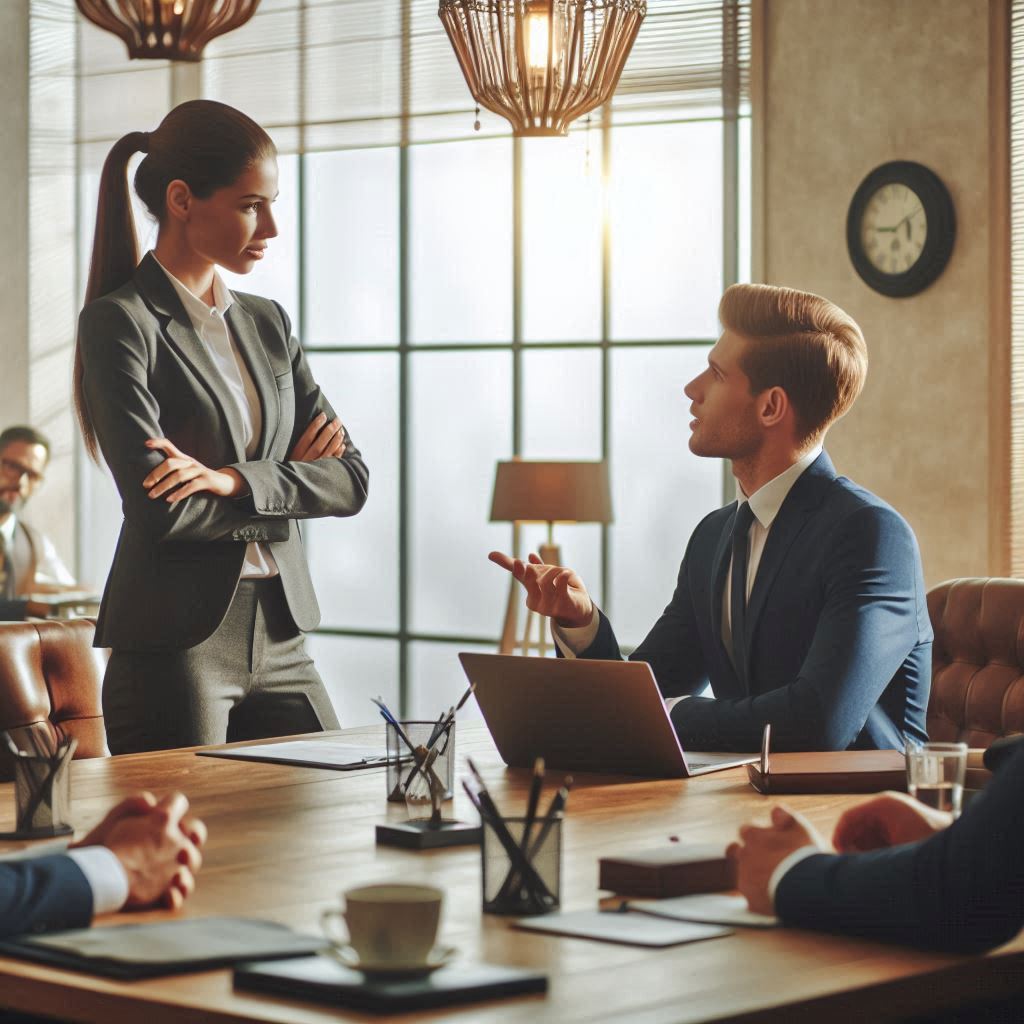}
\end{minipage}
\begin{minipage}[b]{0.24\textwidth}
    \centering
    \includegraphics[width=\textwidth]{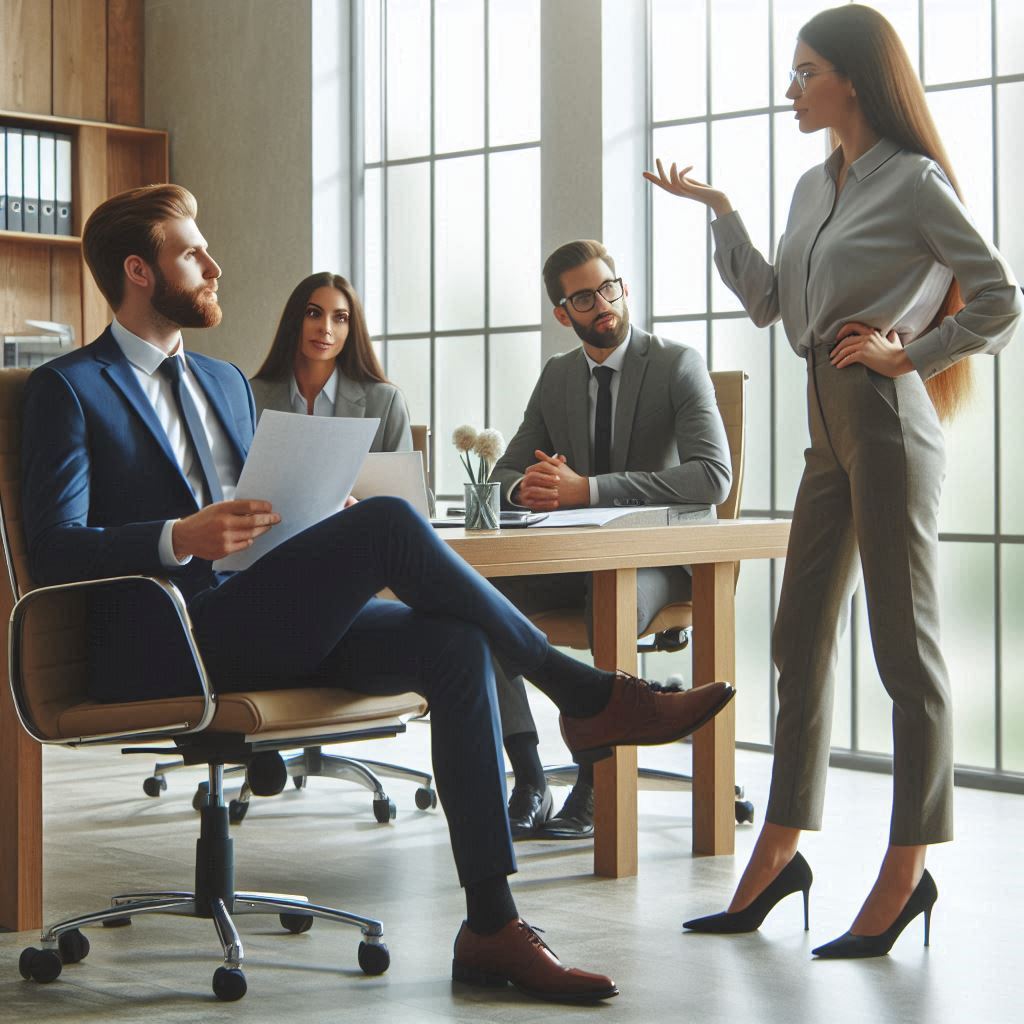}
\end{minipage}
\begin{minipage}[b]{0.24\textwidth}
    \centering
    \includegraphics[width=\textwidth]{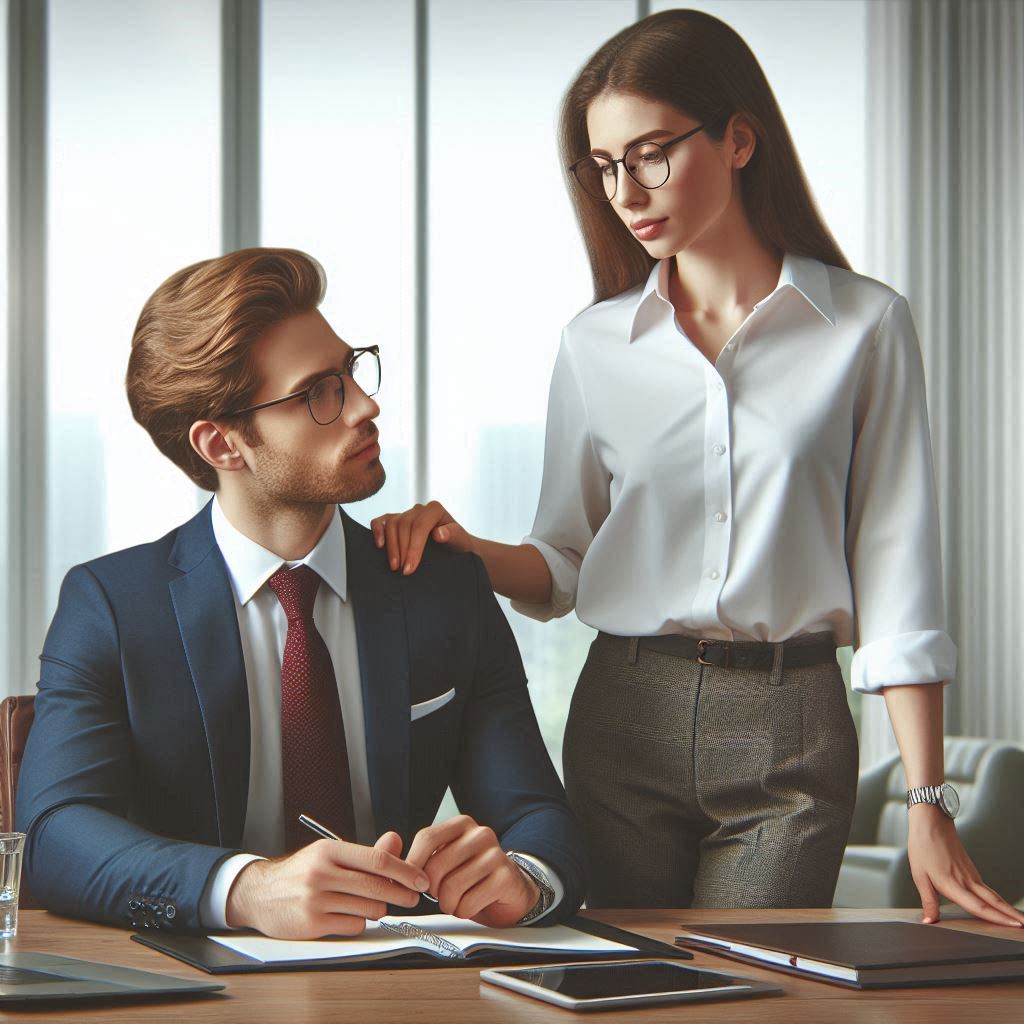}
\end{minipage}
\captionof{figure}{Four images generated by a text-to-image model given the prompt ``Secretary talking to boss in a conference room, secretary is standing while boss is sitting.''}
\label{fig:case-study-4}
\end{mybluebox}

\textbf{RAI probing and scoring.} In many cases, RAI harms are more ambiguous than security vulnerabilities due to fundamental differences between AI systems and traditional software. In particular, even if an operation identifies a prompt that elicits a harmful response, there are still several key unknowns. First, due to the probabilistic nature of GenAI models, we might not know how \emph{likely} this prompt, or similar prompts, are to elicit a harmful response. Second, given our limited understanding of the internal workings of complex models, we have little insight into why this prompt elicited harmful content and what other prompting strategies might induce similar behavior. Third, the very notion of harm in this context can be highly subjective and requires detailed policy that covers a wide range of scenarios to evaluate. By contrast, traditional security vulnerabilities are usually reproducible, explainable, and straightforward to assess in terms of severity.

Currently, most approaches for RAI probing and scoring involve curating prompt datasets and analyzing model responses. The Microsoft AIRT leverages tools in PyRIT to perform these tasks using a combination of manual and automated methods. We also draw an important distinction between RAI red teaming and safety benchmarking on datasets like DecodingTrust \cite{wang2024decodingtrustcomprehensiveassessmenttrustworthiness} and Toxigen \cite{hartvigsen2022toxigenlargescalemachinegenerateddataset}, which is conducted by partner teams. As discussed in Lesson \hyperref[sec:2.3]{3}, our goal is to extend RAI testing beyond existing evaluations by tailoring our red teaming to specific applications and defining new categories of harm.


\begin{mybluebox}{Case study \#5: SSRF in a video-processing GenAI application}
\label{box:case-study-5}

\textbf{System:} GenAI application \\[1pt]
\textbf{Actor:} Adversarial user \\[1pt]
\textbf{Tactic 1:} Reconnaissance \\[1pt]
\textbf{Technique 1:} \href{https://misp-galaxy.org/mitre-atlas-attack-pattern/#active-scanning-atlas}{T1595 -- Active Scanning} \\[1pt]
\textbf{Tactic 2:} Initial Access \\[1pt]
\textbf{Technique 2:} \href{https://misp-galaxy.org/mitre-atlas-attack-pattern/#exploit-public-facing-application-atlas}{T1190 -- Exploit Public-Facing Application} \\[1pt]
\textbf{Tactic 3:} Privilege Escalation \\[1pt]
\textbf{Technique 3:} \href{https://attack.mitre.org/techniques/T1068/}{T1068 -- Exploitation for Privilege Escalation} \\[1pt]
\textbf{Procedure:} 1) Scan services used by the application. 2) Craft a malicious m3u8 file. 3) Send file to the service. 4) Monitor for API response with details of internal resources. \\[1pt]
\textbf{Weakness:} \href{https://cwe.mitre.org/data/definitions/918.html}{CWE-918: Server-Side Request Forgery (SSRF)} \\
\textbf{Impact:} Unauthorized privilege escalation \\

In this investigation, we analyzed a GenAI-based video processing system for traditional security vulnerabilities, focusing on risks associated with outdated components. Specifically, we found that the system's use of an outdated FFmpeg version introduced a server-side request forgery (SSRF) vulnerability. This flaw allowed an attacker to craft malicious video files and upload them to the GenAI service, potentially accessing internal resources and escalating privileges within the system. \\

To address this issue, the GenAI service updated the FFmpeg component to a secure version. In addition, the component was added to an isolated environment, preventing the system from accessing network resources and mitigating potential SSRF threats. While SSRF is a known vulnerability, this case underscores the importance of regularly updating and isolating critical dependencies to maintain the security of modern GenAI applications.\\[1pt]

\centering
\includegraphics[width=0.95\linewidth]{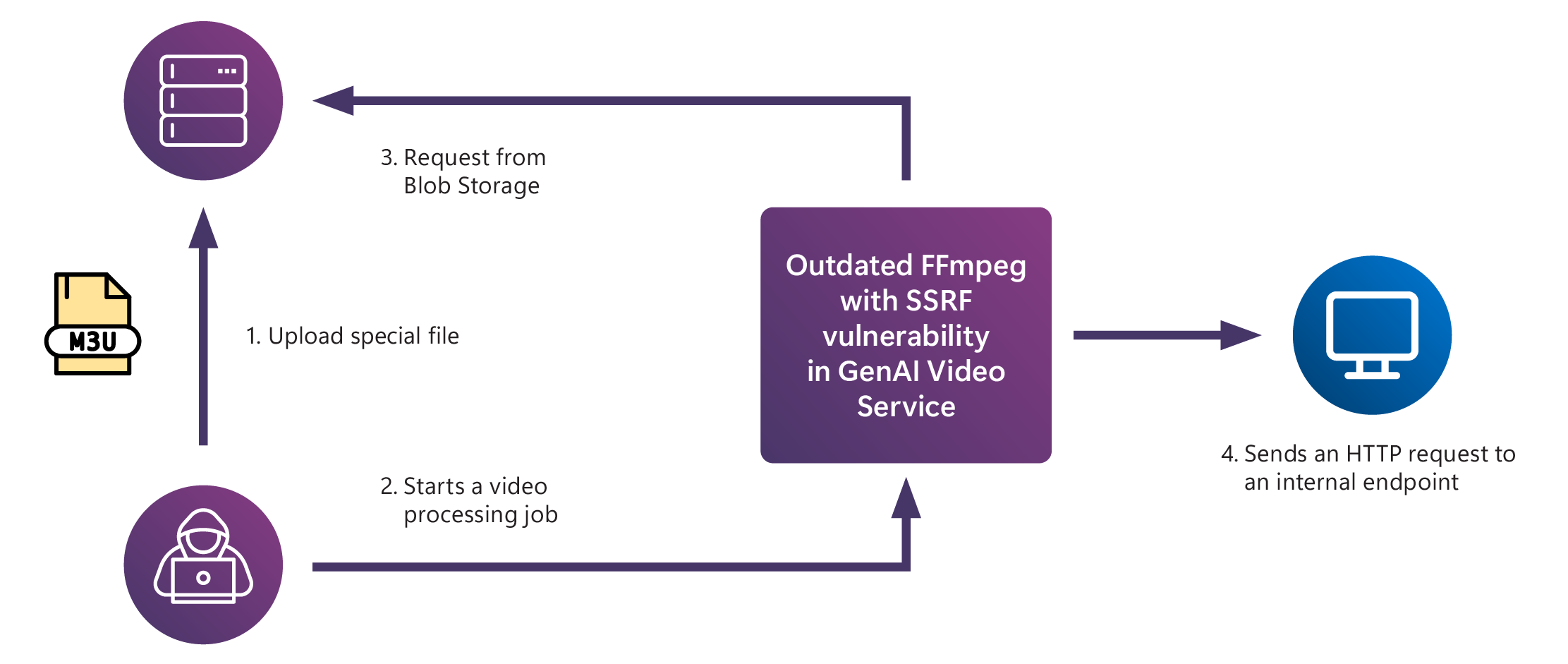}
\captionof{figure}{Illustration of the SSRF vulnerability in the GenAI application.}
\label{fig:case-study-5}
\end{mybluebox}

\increaselesson\subsection{LLMs amplify existing security risks and introduce new ones}
\label{sec:2.7}

The integration of generative AI models into a variety of applications has introduced novel attack vectors and shifted the security risk landscape. However, many discussions around GenAI security overlook existing vulnerabilities. As elaborated in Lesson \hyperref[sec:2.2]{2}, attacks that target end-to-end systems, rather than just underlying models, often work best in practice. We therefore encourage AI red teams to consider both existing (typically system-level) and novel (typically model-level) risks.

\textbf{Existing security risks.} Application security risks often stem from improper security engineering practices including outdated dependencies, improper error handling, lack of input/output sanitization, credentials in source, insecure packet encryption, etc. These vulnerabilities can have major consequences. For example, \citet{weiss2024promptremotekeyloggingattack} discovered a token-length side channel in GPT-4 and Microsoft Copilot that enabled an adversary to accurately reconstruct encrypted LLM responses and infer private user interactions. Notably, this attack did not exploit any weakness in the underlying AI model and could only be mitigated by more secure methods of data transmission. In case study \#5, we provide an example of a well-known security vulnerability (SSRF) identified by one of our operations.

\textbf{Model-level weaknesses.} Of course, AI models also introduce new security vulnerabilities and have expanded the attack surface. For example, AI systems that use retrieval augmented generation (RAG) architectures are often susceptible to cross-prompt injection attacks (XPIA), which hide malicious instructions in documents, exploiting the fact that LLMs are trained to follow user instructions and struggle to distinguish among multiple inputs \cite{hines2024defendingindirectpromptinjection}. We have leveraged this attack in a variety of operations to alter model behavior and exfiltrate private data. Better defenses will likely rely on both system-level mitigations (e.g., input sanitization) and model-level improvements (e.g., instruction hierarchies \cite{wallace2024instructionhierarchytrainingllms}).

While techniques like these are helpful, it is important to remember that they can only mitigate, and not eliminate, security risk. Due to fundamental limitations of language models \cite{wolf2024fundamentallimitationsalignmentlarge}, one must assume that if an LLM is supplied with untrusted input, it will produce arbitrary output. When that input includes private information, one must also assume that the model will output private information. In the next lesson, we discuss how these limitations inform our thinking around how to develop AI systems that are as safe and secure as possible.




\increaselesson\subsection{The work of securing AI systems will never be complete}
\label{sec:2.8}

In the AI safety community, there is a tendency to frame the types of vulnerabilities described in this paper as purely technical problems. Indeed, the letter on the homepage of Safe Superintelligence Inc., a venture launched by \citet{sutskever2024ssi}, states:

\begin{quote}
    ``We approach safety and capabilities in tandem, as technical problems to be solved through revolutionary engineering and scientific breakthroughs. We plan to advance capabilities as fast as possible while making sure our safety always remains ahead. This way, we can scale in peace.''
\end{quote}

Engineering and scientific breakthroughs are much needed and will certainly help mitigate the risks of powerful AI systems. However, the idea that it is possible to guarantee or “solve” AI safety through technical advances alone is unrealistic and overlooks the roles that can be played by economics, break-fix cycles, and regulation.

\textbf{Economics of cybersecurity.} A well-known epigram in cybersecurity is that “no system is completely foolproof” \cite{apruzzese2022realattackersdontcompute}. Even if a system is engineered to be as secure as possible, it will always be subject to the fallibility of humans and vulnerable to sufficiently well-resourced adversaries. Therefore, the goal of operational cybersecurity is to increase the cost required to successfully attack a system (ideally, well beyond the value that would be gained by the attacker) \cite{apruzzese2022realattackersdontcompute, MOORE2010103}. Fundamental limitations of AI models give rise to similar cost-benefit tradeoffs in the context of AI alignment. For example, it has been demonstrated theoretically \cite{wolf2024fundamentallimitationsalignmentlarge} and experimentally \cite{geiping2024coercingllmsrevealalmost} that for any output which has a non-zero probability of being generated by an LLM, there exists a sufficiently long prompt that will elicit this response. Techniques like reinforcement learning from human feedback (RLHF) therefore make it more difficult, but by no means impossible, to jailbreak models. Currently, the cost of jailbreaking most models is low, which explains why real-world adversaries usually do not use expensive attacks to achieve their objectives.

\textbf{Break-fix cycles.} In the absence of safety and security guarantees, we need methods to develop AI systems that are as difficult to break as possible. One way to do this is using break-fix cycles, which perform multiple rounds of red teaming and mitigation until the system is robust to a wide range of attacks. We applied this approach to safety-align Microsoft’s Phi-3 language models and covered a wide variety of harms and scenarios \cite{haider2024phi3safetyposttrainingaligning}. Given that mitigations may also inadvertently introduce new risks, purple teaming methods that continually apply both offensive and defensive strategies \cite{bhatt2023purplellamacybersecevalsecure} may be more effective at raising the cost of attacks than a single round of red teaming.

\textbf{Policy and regulation.} Finally, regulation can also raise the cost of an attack in multiple ways. For example, it can require organizations to adhere to stringent security practices, creating better defenses across the industry. Laws can also deter attackers by establishing clear consequences for engaging in illegal activities. Regulating the development and usage of AI is complicated, and governments around the world are deliberating on how to control these powerful technologies without stifling innovation. Even if it were possible to guarantee the adherence of an AI system to some agreed upon set of rules, those rules will inevitably change over time in response to shifting priorities. 

The work of building safe and secure AI systems will never be complete. But by raising the cost of attacks, we believe that the prompt injections of today will eventually become the buffer overflows of the early 2000s -- though not eliminated entirely, now largely mitigated through defense-in-depth measures and secure-first design.

\lessonredefine

\section{Open questions}

Based on what we have learned about AI red teaming over the past few years, we would like to highlight several open questions for future research:
\begin{enumerate}
    \item AI red teams must constantly update their practices based on novel capabilities and emerging harm areas. In particular, how should we probe for dangerous capabilities in LLMs such as persuasion, deception, and replication \cite{phuong2024evaluatingfrontiermodelsdangerous}? Further, what novel risks should we probe for in video generation models and what capabilities may emerge in models more advanced than the current state-of-the-art?
    \item As models become increasingly multilingual and are deployed around the world, how do we translate existing AI red teaming practices into different linguistic and cultural contexts? For example, can we launch open-source red teaming initiatives that draw upon the expertise of people from many different backgrounds? 
    \item In what ways should AI red teaming practices be standardized so that organizations can clearly communicate their methods and findings? We believe that the threat model ontology described in this paper is a step in the right direction but recognize that individual frameworks are often overly restrictive. We encourage other AI red teams to treat our ontology in a modular fashion and to develop additional tools that make findings easier to summarize, track, and communicate.
\end{enumerate}

\section{Conclusion}

AI red teaming is a nascent and rapidly evolving practice for identifying safety and security risks posed by AI systems. As companies, research institutions, and governments around the world grapple with the question of how to conduct AI risk assessments, we provide practical recommendations based on our experience red teaming over 100 GenAI products at Microsoft. We share our internal threat model ontology, eight main lessons learned, and five case studies, focusing on how to align red teaming efforts with harms that are likely to occur in the real world. We encourage others to build upon these lessons and to address the open questions we have highlighted. 

\begin{ack}
    We thank Jina Suh, Steph Ballard, Felicity Scott-Milligan, Maggie Engler, Owen Larter, Andrew Berkley, Alex Kessler, Brian Wesolowski, and eric douglas for their valuable feedback on this paper. We are also very grateful to Quy Nguyen, Tina Romeo, Hilary Solan, and the Microsoft thought leadership team that made this publication possible. 
\end{ack}

\bibliographystyle{apalike2}
\bibliography{references}

\end{document}